\documentclass{article}

\usepackage{arxiv}

\usepackage[utf8]{inputenc} 
\usepackage[T1]{fontenc}    
\usepackage{hyperref}       

\usepackage{url}            
\usepackage{booktabs}       
\usepackage{amsfonts}       
\usepackage{nicefrac}       
\usepackage{microtype}      
\usepackage{lipsum}
\usepackage{graphicx}
\usepackage{multirow}
\usepackage[table,xcdraw]{xcolor}
\graphicspath{ {./images/} }
\usepackage{pifont}
\usepackage{amsmath} 
\usepackage{cleveref} 

\usepackage{etoolbox} 
\usepackage{subcaption}
\usepackage{algorithm}
\usepackage{algorithmic}

\title{ZeroSense:How Vision matters in Long Context Compression}

\author{
  Yonghan Gao\textsuperscript{1}$^\dagger$,
  Zehong Chen\textsuperscript{1}$^\dagger$,
  Lijian Xu\textsuperscript{1}$^*$,
  Jingzhi Chen\textsuperscript{1},
  Jingwei Guan\textsuperscript{2},
  Xingyu Zeng\textsuperscript{1}$^*$ \\
  \textsuperscript{1}Shenzhen University of Advanced Technology,
  \textsuperscript{2}Shenzhen Technology University \\
  $^\dagger$Equal contribution \\
  $^*$Corresponding authors: \texttt{\{xulijian, zengxingyu\}@suat-sz.edu.cn}
}

\begin{document}
\maketitle

\begin{abstract}
Recent visual-text compression (VTC) methods, typified by DeepSeek-OCR, report impressive high token compression ratios for long-context modeling tasks by leveraging text-to-image rendering. However, existing evaluation protocols heavily rely on downstream task performance. Such evaluation metrics fail to accurately measure text preservation due to the strong inherent linguistic priors of Multimodal Large Language Models (MLLMs). In this work, we introduce a new evaluation framework that decouples MLLMs' capabilities to faithfully assess VTC quality. Within this framework, we further introduce the ZeroSense Benchmark to ensure low semantic correlation of testing samples. By eliminating contextual dependencies, our benchmark guarantees that the evaluation results are purely reflective of VTC quality, unaffected by the semantic inference capabilities of downstream models. Extensive experiments across multiple datasets demonstrate that VTC quality and downstream task accuracy diverge significantly, highlighting the necessity of our decoupled evaluation framework. Our code is available at \url{https://github.com/MedHK23/ZeroSense}.
\end{abstract}

\section{Introduction}
The escalating demand for long-context processing in Large Language Models (LLMs) has encountered a fundamental bottleneck due to the quadratic complexity of self-attention. 
Distinct from traditional compression methods for long-context modeling~\cite{chen2023longlora,chen2025minimax,chen2024long}, the Visual-Text Compression (VTC) paradigm proposes rendering extensive textual sequences into compact document images, effectively substituting thousands of textual tokens with visual tokens. 
Beyond DeepSeek-OCR~\cite{wei2025deepseek}, systems such as Glyph~\cite{cheng2025glyph} and VTC-R1~\cite{wang2026vtc} have demonstrated significant potential in inference acceleration and context extension.

\begin{figure}[t]
  \centering
  \includegraphics[width=1\linewidth]{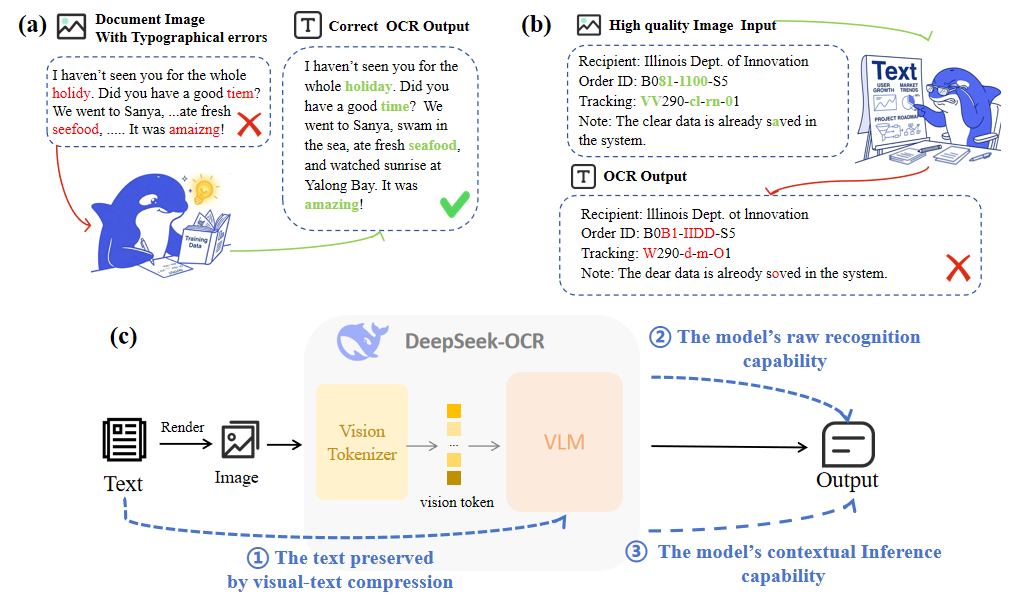}
  
  \caption{\textbf{Decoupled analysis of confounding factors in visual-text compression.} 
  \textbf{(a) Semantic Priors Compensation:} The model leverages contextual inference to rectify typographical errors in the source image. 
  \textbf{(b) Raw Perceptual Bottleneck:} In a semantic vacuum, the model exhibits failures on incoherent alphanumeric strings despite high input resolution, revealing the intrinsic ceiling of its raw recognition capability. 
  \textbf{(c) Performance Decomposition Framework:} End-to-end performance is deconstructed into the synergistic interaction between The text preserved by visual-text compression, The model’s raw recognition capability, and the model’s contextual inference capability.}
    \label{fig:framework}
    \vspace{-10pt}
\end{figure}

Our empirical investigation reveals a complex coupling between the text preservation of the specific VTC strategy and downstream task performance. As illustrated in~\Cref{fig:framework}(a), DeepSeek-OCR exhibits robust contextual inference by autonomously correcting typographical errors within the source image. While it enhances robustness, it simultaneously obscures the text loss incurred during compression. Recent studies confirm that end-to-end OCR accuracy degrades catastrophically when the input's semantic structure is compromised~\cite{liang2026visual}. Furthermore,~\Cref{fig:framework}(b) shows that even with high-resolution inputs, documents with weak semantic correlations often yield "glyph-similar" erroneous outputs. This indicates that the model's performance on downstream tasks cannot be simply interpreted as the quality of visual-text compression.

\section{Related Work}
\subsection{Evolution of Document Understanding: From OCR to Optical Compression}
\label{sec:related_compression}
Document understanding models face a fundamental trade-off between representation fidelity and computational efficiency. Early methods followed a cascade paradigm, relying on OCR engines to extract text for processing by language models~\cite{smith2007overview}. Although effective for text-dominant documents, these approaches discard layout cues and suffer from error propagation~\cite{xu2020layoutlm}.

Driven by the Transformer architecture, document understanding has shifted toward unified end-to-end frameworks.
Representative directions include encoder–decoder OCR models~\cite{li2023trocr} and document intelligence systems that jointly model textual content, layout structure, and visual appearance~\cite{xu2020layoutlm,tang2023unifying}.
Concurrently, OCR-free paradigms have gained traction, bypassing text recognition by directly predicting structured outputs from document images or leveraging large-scale image-to-text pretraining for visually grounded language modeling~\cite{kim2022ocr,lee2023pix2struct,young2026scalar}.
These end-to-end pixel-based models address the limitations of pipeline OCR systems by directly mapping image pixels to output sequences while preserving layout semantics.
However, they struggle with token efficiency in large-scale visual representation learning~\cite{yang2025one}.
 Encoding high-resolution document images often requires thousands of visual tokens, making them computationally expensive for long-context applications~\cite{blecher2023nougat,young2026fewer}.

Several studies mitigate this scalability issue through compression. Methods such as DeepSeek-OCR~\cite{wei2025deepseek}, Context Cascade Compression~\cite{liu2025context} and VTC-R1~\cite{wang2026vtc} use downsampling or perceiver resamplers to compress a full document page into a fixed number of tokens. While these techniques improve efficiency, they obscure how much text is preserved rather than reconstructed from the semantic priors. Current evaluation protocols rely on question-answering tasks where linguistic priors can conceal the text loss caused by compression. Consequently, a framework that quantifies the text retained by compression modules in a semantics-free setting remains needed. This work introduces such a framework.

\subsection{Benchmarks for Document Analysis and Compression Evaluation}
\label{sec:related_benchmarks}

Existing benchmarks for document analysis\cite{ma2024mmlongbench,bai2024longbench,cheng2023m6doc,pfitzmann2022doclaynet} and compression evaluation broadly fall into two categories, both of which exhibit limitations when assessing optical compression.
The first category encompasses foundational optical character recognition datasets, including TextOCR~\cite{singh2021textocr}, OCRbench~\cite{liu2024ocrbench}, the ICDAR series~\cite{karatzas2015icdar}, etc. While these benchmarks evaluate the fundamental perception capability of extracting text, they present two major limitations for high-ratio optical compression. They often exhibit a scene-text bias by featuring sparse text with low density, which fails to adequately test compression limits. Furthermore, they typically rely on fragmented bounding boxes for isolated words. This approach lacks the complex, high-density global layouts found in real-world documents, such as multi-column text or dense tables. Consequently, these datasets cannot assess whether a compression module preserves a document's macroscopic structure.

Although these benchmarks assess end-to-end task performance, they are more importantly affected by semantic priors. The strong semantic coherence in these documents enables language models to guess missing text using linguistic priors, which masks the visual distortion caused by extreme compression.
To evaluate long-context document reasoning and compression, researchers typically rely on visual document understanding benchmarks such as Fox~\cite{liu2024focus}, Omni~\cite{ouyang2025omnidocbench} and DocVQA~\cite{mathew2021docvqa}. Although these benchmarks assess end-to-end task performance, they are affected by semantic priors. The strong semantic coherence in these documents enables language models to guess missing text using linguistic priors, which masks the visual distortion caused by extreme compression. While some recent studies introduce text perturbation or shuffling techniques~\cite{chai2024tokenization} to test visual grounding, and others propose compression-oriented long-context evaluations such as VTCBench~\cite{zhao2025vtcbench}, which measures retrieval, associative reasoning, and dialogue memory under visual-text compression settings, the testing samples in these benchmarks are still semantically  correlated. As a result, models may exploit residual semantic priors rather than purely visual signals. Such data-level interventions either disrupt natural document layouts or fail to eliminate latent linguistic dependencies. Consequently, they fail to completely decouple visual signals from linguistic priors.

These limitations highlight the need for a benchmark that maintains realistic document layouts while eliminating semantic priors. The ZeroSense benchmark addresses this requirement. By establishing a semantic vacuum that preserves the structural complexity of full-page documents, it provides a controlled setting to evaluate the optical fidelity of compression models without the confounding effect of linguistic priors.

\section{Framework}
\label{sec:framework}
This section presents an evaluation framework for assessing different VTC strategies. \Cref{sec:Formulation} formalizes the VTC task, defines the compression ratio, and establishes evaluation criteria bounded by expectation and model capabilities. \Cref{sec:Formulation_Task} maps this theoretical formulation to a practical OCR task. To isolate the model's text preservation from semantic priors and varying base OCR proficiencies, we introduce a decoupled modeling framework (see \Cref{sec:Decoupling}). This approach provides a rigorous methodology for measuring text loss during visual rendering.
\subsection{Formulation}
\label{sec:Formulation}
When handling long-context tasks, the model mainly involves three variables: the user-specified task $I$, the long input text $C = \{c_1, \dots, c_t\}$ to be processed, and the model’s prediction output $O$.
The core optimization objective in this setting is to maximize:
\begin{equation}
F(O \mid I, C)
\end{equation}
Although the length of $I$ is limited and can generally be neglected, the length of $C$ often becomes excessively long, which imposes a heavy burden on the model’s computational power and memory.
Recent studies have shown that by VTC, text $C$ can be rendered in an image sequence $V_\theta = \{v_1, \dots, v_n\}$, effectively reducing the number of input tokens.In this case, the optimization objective is changed to:
\begin{equation}
    F (O | I, V_\theta)
\end{equation}
To quantify the spatial efficiency of a specific rendering mapping $\theta$, we define the compression ratio $\rho(\theta)$ as the quotient of the original text sequence length and the aggregate number of visual tokens $\tau(v_i)$ across all generated frames:
\begin{equation}
    \rho(\theta) = \frac{|C|}{\sum_{i=1}^n \tau(v_i)}
\end{equation}
This metric explicitly captures the number of original textual tokens encoded per visual token.
Our primary contribution is the formalization of a robust evaluation protocol to quantify the performance discrepancy between visually compressed and pure-text inputs across varying compression ratios $\rho(\theta)$. To ensure theoretical validity, this protocol enforces two critical constraints. First, the measured performance delta must be invariant to the semantic content of $C$. Because the rendering mapping $\theta$ must generalize across arbitrary inputs, we compute the expected value over the input distribution $\mathcal{D}$. Second, the absolute inference likelihood $F$ is fundamentally bounded by the OCR capabilities of the evaluated MLLM. Suboptimal empirical results may reflect an inherent visual encoding deficit rather than algorithmic limitations in $\theta$. To isolate the theoretical upper bound of the compression strategy, we evaluate a predefined set of models $\mathcal{M}$ and extract the maximum achieved performance. Integrating these constraints yields the unified evaluation metric:
\begin{equation}
    \mathcal{S}(\theta) = \max_{m \in \mathcal{M}} \mathbb{E}_{C \sim \mathcal{D}} \left[ F_m(O \mid I, V_\theta) - F_m(O \mid I, C) \right]
\end{equation}
If $\mathcal{S}(\theta)$ converges toward zero for at least one candidate model $m \in \mathcal{M}$, the rendering strategy effectively preserves the essential semantic information at the specified compression ratio.
\subsection{OCR Formulation Task}
\label{sec:Formulation_Task}
To measure the text loss caused by VTC during evaluation, we align with DeepSeek-OCR by mapping the downstream task to OCR text restoration. Here, the target response $O$ becomes the original text $C$. Substituting this into our previous formulation shifts our focus to the correlation between $\rho(\theta)$ and $F(C \mid I, V_\theta) - F(C \mid I, C)$. 
Since the probability of generating $C$ given the text $C$ is theoretically 100\%, the evaluation metric simplifies to the relationship between the compression ratio $\rho(\theta)$ and $F(C \mid I, V_\theta)$. 
However, directly using $F(C \mid I, V_\theta)$ introduces two limitations. First, mainstream OCR benchmarks such as FOX and Omni consist primarily of everyday documents with strong contextual correlations. Large language models can use semantic priors to infer correct words from context even when visual information is degraded, artificially inflating $F(C \mid I, V_\theta)$. Second, as noted earlier, a foundation model with weaker base recognition capabilities yields a lower $F(C \mid I, V_\theta)$, failing to accurately reflect the rendering method's  text preservation. To isolate these confounding factors and quantify text preservation, we propose a decoupled model for the end-to-end text generation probability. The conditional generation probability for any text segment or token $C_i$ in the sequence can be decomposed into logical reasoning derived from historical context and text extraction from the current visual page:
\begin{equation}
    F(C_i \mid I, V_\theta) \approx F_{prior}(C_i \mid I, V_{\le i}) + OCR_{raw}(C_i \mid V_i) \cdot K_{quality}
\end{equation}
Within this decoupled framework, $F_{prior}(C_i \mid I, V_{\le i})$ denotes the probability of a prior guess driven by contextual semantic correlations, and $OCR_{raw}(C_i \mid V_i)$ captures the model's inherent capability to extract characters from a single visual page segment. Crucially, $K_{quality}$ is defined as the text preservation of the specific VTC strategy. This formulation clarifies that the end-to-end performance is a joint result of the model's prior reasoning capability, its raw recognition capability, and the text preserved by the rendering method. By utilizing benchmarks such as ZeroSense to create a semantic vacuum, we can explicitly isolate and approximate $F_{prior}$. Concurrently, $OCR_{raw}$ can be independently measured. This allows us to rearrange the equation to derive the text preserving rate:
\begin{equation}
    K_{quality} \approx \frac{F(C \mid I, V_\theta) - F_{prior}(C \mid I, V)}{OCR_{raw}(C \mid V)}
\end{equation}
This derivation enables us to authentically evaluate the text preservation capability of visual representations across different compression strategies, independent of the base model's reasoning bias or raw recognition bounds.

\subsection{Decoupling Semantic Priors and OCR}
\label{sec:Decoupling}
In order to measure $F_{prior}$, we introduce a ZeroSense benchmark consisting of non-semantic-correlation character sequences. By maintaining identical rendering parameters $\theta$, we can quantify $F_{prior}$ as 
\begin{equation}
    F_{prior} = F(C|I,V_{\theta}) -  F(C|I, V_{\theta}(ZeroSense))
\end{equation}
where $F(C|I, V_{\theta}(ZeroSense))$ represents the downstream task performance on the ZeroSense Benchmark. Details of the ZeroSense Benchmark can be found in the following ~\Cref{sec:benchmark}.

Meanwhile, a calibrated accuracy baseline is proposed to account for the inherent OCR perceptual ability $OCR_{raw}(C|V)$ of Multimodal Large Language Models (MLLMs). We hypothesize that the model's OCR performance exhibits a systematic functional dependency on the intensity of visual-text compression. Specifically, we posit that the degradation in OCR accuracy can be modeled as a linear function of the compression ratio, allowing for a predictable mapping between visual quality and recognition fidelity. To empirically derive this baseline, we curate a subset  of high-fidelity reference samples verified by human annotators to ensure the text is fully preserved during rendering. By evaluating the MLLMs on these samples, we can isolate the model's inherent OCR capability from the artifacts introduced by the compression. 

\begin{figure}[!b]
  \centering
  \includegraphics[width=1\linewidth]{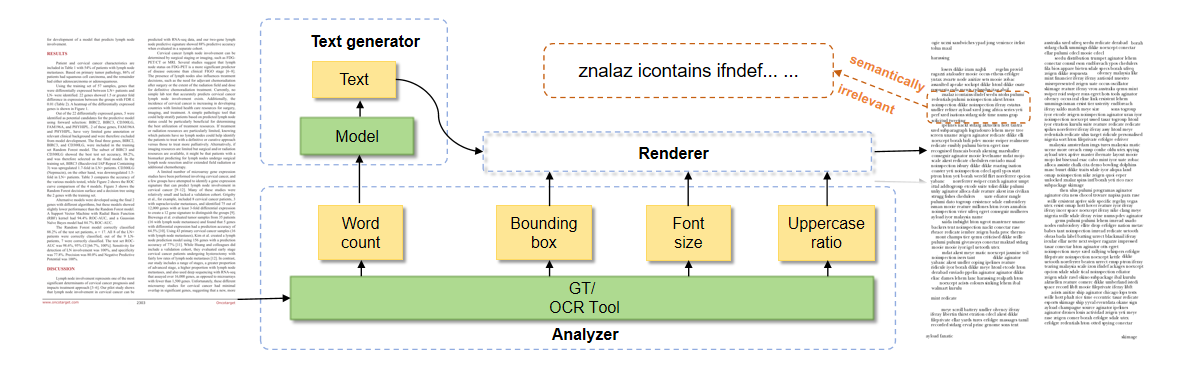} 
  \caption{Pipeline for semantically irrelevant text generation and rendering. 
The system comprises three modules: an Analyzer that extracts key visual attributes 
(e.g., word counts and bounding-box coordinates) from input images using OCR Tool
or ground-truth annotations; a Text Generator that produces semantically irrelevant 
text with LLM; and a Renderer that synthesizes the final images by 
combining the extracted visual attributes with the generated text.}

  \label{fig:benchmark_flow}
\end{figure}

\section{ZeroSense Benchmark}
\label{sec:benchmark}

Current benchmarks for OCR tasks are predominantly sourced from daily documents. Consequently, the textual content within these datasets exhibits strong semantic correlations. To prevent these semantic priors from biasing our measurement of the model's text preservation capability, we propose the construction of an auxiliary dataset, termed the ZeroSense dataset. This auxiliary benchmark is designed to provide the precise rendering layouts $\theta$ (font size, case, capacity, line height, etc.) required by specific methods, while strictly guaranteeing the semantic irrelevance of the visual content. We herein outline the overall dataset construction pipeline (Section~\ref{sec:benchmark_pipline}), detail the extraction of layout features (Section~\ref{sec:Feature_Extraction}), explain the algorithm for ensuring semantic independence (Section~\ref{sec:Semantically irrelevant sample generation}). The dataset statistics are presented in the appendix.

\subsection{Pipeline}
\label{sec:benchmark_pipline}
When evaluating a VTC method, the ideal approach uses the method's explicitly defined rendering configuration. However, leading baselines like DeepSeek-OCR directly use datasets such as Fox and Omni, implicitly adopting their inherent document rendering styles. To ensure a fair comparison with these baselines, we reverse-engineer and extract these specific layout configurations. If our framework evaluates other methods that explicitly provide their configurations, this extraction step can be bypassed. The construction pipeline consists of three stages. We first extract the layout features necessary for subsequent rendering from the source datasets. Next, we use the posterior probabilities of a language model to generate semantically agnostic textual samples, replacing the original document content. Finally, we render the visual pages by combining the extracted spatial features with the generated zero-semantics text. Using the extracted masks, we apply an inpainting algorithm to remove the original text while preserving background textures and noise patterns.

\subsection{Feature Extraction}
\label{sec:Feature_Extraction}
To precisely replicate the rendering styles of the baselines, we extract the necessary configuration parameters $\theta$(font size $S_{opt}$, bounding box $B_{pos}$, the text capacity $C_i$..) based on the original annotations provided by Fox and Omni. The inputs for this extraction are the bounding box coordinates $B_{pos} = (x, y, w, h)$ and the original text content $T$.
The extraction algorithm for each layout feature is formulated as follows:

\paragraph{\textbf{Bounding box ($B_{pos}$)}}
Some datasets directly provide render-friendly paragraph-level bounding box coordinates, which can be directly reused. However, the ground truth in many datasets lacks precise positional information for paragraph, or provides highly scattered bounding boxes without paragraph-level structure. To recover the document layout, we propose a bottom-up reconstruction algorithm based on vertical projection and neighborhood aggregation. We use vertical projection profiles to identify column gaps, segmenting the page into independent columns. Within each column, we apply heuristic aggregation rules based on geometric distance: adjacent text boxes are merged if their vertical projections overlap and their vertical distance falls below a predefined threshold. This process integrates scattered word-level bounding boxes into cohesive paragraph-level text blocks.

\paragraph{\textbf{Font size ($S_{opt}$)}}
Estimating typographic attributes, particularly font size, is critical for recognition performance, yet this information is rarely available from raw OCR results. To address this, we design a geometry-constrained inverse solver. For a given text block $T$ and its bounding box dimensions $(w, h)$, we define the optimal font size $S_{opt}$ as the maximum integer value that allows the rendered text to occupy the region. We construct a rendering simulation function $f(T, s)$ to calculate the physical dimensions of the text at a given font size $s$. We formalize this objective as:
\begin{equation}
    S_{opt} = \max \{ s \in \mathbb{N} \mid h(f(T, s)) \le h \land w(f(T, s)) \approx w \}
    \label{eq:max}
\end{equation}
To solve this, we use an iterative search strategy to adjust the font size. We also introduce a language-specific buffer coefficient to compensate for rendering engine discrepancies, which helps recover the visual hierarchy of the document.

\paragraph{\textbf{Text Capacity ($C_i$)}}
To generate sufficient semantically irrelevant text in subsequent stages, we calculate the capacity of each text block. Because of encoding differences between languages, we use a language-aware counting strategy. For Latin-based scripts, such as English, capacity is measured in words. For logographic systems, such as Chinese, it is measured in characters. 
Formally, the capacity $C_i$ of a text block $B_i$ containing text $T_i$ is defined as:
\begin{equation}
    C_i = \begin{cases} count_{word}(T_i), & \text{if } L(T_i) = Latin \\ count_{char}(T_i), & \text{if } L(T_i) = Logographic \end{cases}
    \label{eq:count}
\end{equation}
where $L(\cdot)$ denotes a language detection function based on ASCII density. This capacity metric complements the previously defined typographic attributes to describe the document's visual-semantic density.

\subsection{Semantically Irrelevant Sample Generation}
\label{sec:Semantically irrelevant sample generation}
Because text sequences in existing datasets exhibit strong semantic dependencies, MLLMs often bypass visual decoding and rely on semantic priors to infer missing tokens. To counter this, we use a replacement algorithm driven by the posterior probabilities of a causal language model to generate contextually irrelevant content. For a given context $x_{<t}$, we force the model to sample the next token $w$ from a restricted vocabulary subset $Q_t$, defined by a low-probability threshold $\tau$:
\begin{equation}
    Q_t = \{ w \in V_{valid} \mid P(w \mid x_{<t}) < \tau \}
    \label{eq:valid}
\end{equation}
where $V_{valid}$ is the valid vocabulary space. By replacing the original text $T$ with tokens sampled from $K_t$, we disrupt contextual predictability. This creates a semantic vacuum that forces the evaluation to rely entirely on the visual retention of the rendered page rather than the language model's prior knowledge.

\section{Experiment}
\label{sec:Experiment}
\subsection{Experimental Setup}\label{sec:setup}
\noindent\textbf{Models.}
We use the DeepSeek-OCR architecture, deployed via \texttt{vLLM} on a single NVIDIA RTX A6000 GPU. Unless otherwise specified, we adopt the official hyperparameters and decoding strategies for fair comparison.
Our image preprocessing differs from the original DeepSeek-OCR pipeline, which uses mixed resizing that stretches images smaller than $1024 \times 1024$ and pads larger ones. Stretching distorts aspect ratios and glyph shapes, introducing an uncontrolled variable in optical fidelity evaluation. To avoid this distortion, we pad all input document images onto a blank $1280 \times 1280$ canvas according to different
resolution. This aspect-ratio-preserving padding ensures the visual encoder processes the original typography and layout scale.
\begin{table}[htbp]
  \caption{Text preservation metrics across the Fox and Omni datasets at varying compression ratios. The table contrasts the accuracy of the DeepSeek-OCR with $K_{quality}$ computed via our decoupled framework. Results highlight the difference between the two frameworks.}
  \label{tab:comparison_results}
  \centering
  \renewcommand{\arraystretch}{1.2} 
  \begin{tabular}{@{} c c@{\hspace{5pt}}c @{\hspace{14pt}}c@{\hspace{5pt}}c @{}} 
    \toprule
    \multirow{2}{*}{Compression} & \multicolumn{2}{c}{Omni} & \multicolumn{2}{c}{Fox} \\
    \cmidrule(lr){2-3} \cmidrule(lr){4-5} 
          & Our framework & DeepSeek-OCR & Our framework & DeepSeek-OCR \\
    \midrule
    7.5$\times$  & 97.1\% & 89.2\% & 93.9\% & 95.5\% \\
    10$\times$   & 91.5\% & 86.4\% & 77.2\% & 93.4\% \\
    12.5$\times$ & 81.5\% & 84.2\% & 57.1\% & 90.6\% \\
    15$\times$   & 67.5\% & 78.8\% & 38.9\% & 83\%   \\
    17.5$\times$ & 55.7\% & 67.4\% & 27.4\% & 81.3\% \\
    \bottomrule
  \end{tabular}
  \vspace{-2pt}
\end{table}

\noindent\textbf{Benchmarks \& Metrics.} We evaluate the model on four datasets. To assess document understanding and end-to-end performance, we use three benchmarks: Fox, Omni, and DI-100 v1.3. Furthermore, the ZeroSense dataset underpins our evaluation framework and enables the separation of our theoretical variables. 

We evaluate generation quality and operationalize our theoretical framework using two metrics. To measure $F_{prior}(C_i \mid I, V_{\le i})$, we apply character-level precision and normalized edit distance. Character-level precision evaluates the accuracy of generated textual tokens against the ground truth without penalizing minor formatting variations. Normalized edit distance, derived from the Levenshtein distance, calculates the minimum operations required to match the target string. We report this distance on a 0 to 1 scale, where higher values indicate greater structural and character-level alignment. For the subsequent quantitative isolation of $K_{quality}$ and $OCR_{raw}$, we restrict our measurement to character-level precision to maintain a consistent evaluation baseline.

\begin{figure}
    \centering
    \includegraphics[width=1\linewidth]{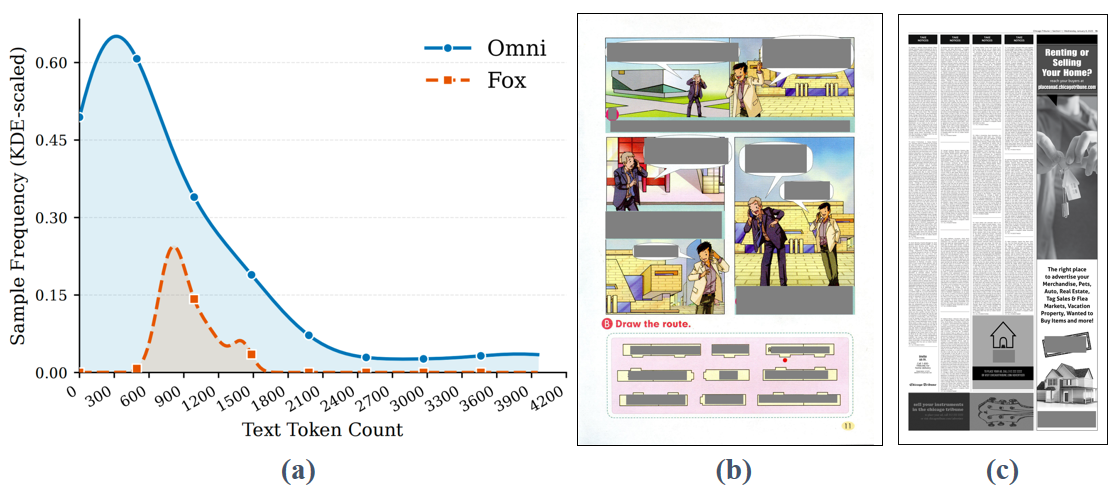}
    \caption{(a) The text token distribution density of the Fox and Omni datasets. 
    The data distribution for Fox is highly concentrated around 900 tokens. Conversely, Omni exhibits a prominent long-tail characteristic; a substantial volume of its data is clustered between 300 and 600 tokens, yet the proportion exceeding 2,100 tokens remains non-negligible. (b) and (c) illustrate typical document examples with extremely low and high token counts, respectively.}
    \label{fig:example}
    \vspace{-15pt}
\end{figure}

\subsection{The Text Preserving Ability of Visual-Text Compression}
\label{subsec:main_results}
Table~\ref{tab:comparison_results} delineates the isolated text preservation, computed via our decoupled framework, against the DeepSeek-OCR paradigm. An observation from these results is the divergence in metric behavior across the two datasets. Specifically, our proposed metric yields lower retention scores than the baseline on the Fox dataset, yet reports higher retention on the Omni dataset. This dichotomy supports our decoupled formulation, exposing two confounding factors in standard evaluations. On the Fox dataset, the standard end-to-end evaluation exhibits higher performance, registering a 81.3\% accuracy at a $17.5\times$ compression ratio. By contrast, our framework indicates a degradation in visual information, with $K_{quality}$ decreasing to 27.4\%. Because the Fox dataset comprises documents with contextual dependencies, the MLLM leverages its semantic priors to infer tokens despite visual degradation. Consequently, traditional end-to-end metrics fail to penalize visual information loss. Our framework disentangles this semantic compensation, revealing the degradation of visual fidelity.
Conversely, on the Omni dataset, our computed $K_{quality}$ (e.g., 97.1\% at $7.5\times$ compression) exceeds the standard end-to-end accuracy (89.2\%).

As shown in Table~\ref{tab:comparison_results}, we observe a significant phenomenon where the performance discrepancy between our framework and DeepSeek-OCR on the Fox dataset ($53.9\%$) is substantially larger than that on the Omni dataset ($11.7\%$) when the compression ratio reaches an extreme of $17.5\times$. A detailed investigation reveals that this phenomenon primarily stems from the distinct distributions of text tokens between the Omni and Fox datasets as illustrated in~\Cref{fig:example}. Specifically, the sample text density in the Fox dataset is relatively uniform, resulting in a concentrated distribution of text tokens. Conversely, the Omni dataset contains a significant number of samples across various text token intervals. For images with low text token counts, the text density is minimal and the semantic information is sparse, making the characters clearly legible. In contrast, images with high text token counts exhibit extreme text density, rendering the characters nearly unrecognizable. In both scenarios, the OCR output remains largely unaffected by the inherent capability of the model to extract characters or its contextual inference capability, which leads to a negligible performance gap between our framework and DeepSeek-OCR. Our experimental setup involves various resolution scaling methods, which results in the inclusion of numerous samples with high text density in the Omni dataset under high compression ratios, thereby reducing the overall average performance difference.

\begin{table}[t]
  \caption{Quantification of semantic priors reliance ($F_{prior}$) under increasing compression ratios. The values indicate the proportion of predictions driven by semantic context rather than visual features. Data is reported for both the Fox and Omni datasets across multiple compression intervals.} 
  \label{tab:compression_results}
  \centering
  \renewcommand{\arraystretch}{1.2} 
  \setlength{\tabcolsep}{6pt} 
  \begin{tabular}{@{}ccccccc@{}}
    \toprule
    \multirow{2}{*}{Compression} & \multicolumn{2}{c}{$F_{prior} + OCR_{raw} \cdot K_{quality}$} & \multicolumn{2}{c}{$OCR_{raw} \cdot K_{quality}$} & \multicolumn{2}{c}{$F_{prior}$} \\
    \cmidrule(lr){2-3} \cmidrule(lr){4-5} \cmidrule(lr){6-7} 
          & Omni   & Fox    & Omni   & Fox    & Omni                   & Fox                   \\
    \midrule
    7.5$\times$  & 77.4\% & 95.5\% & 40.4\% & 71.7\% & \textcolor[HTML]{C00000}{37\%}   & \textcolor[HTML]{C00000}{23.8\%} \\
    10$\times$   & 75.7\% & 93.4\% & 30.4\% & 51.9\% & \textcolor[HTML]{C00000}{45.3\%} & \textcolor[HTML]{C00000}{41.5\%} \\
    12.5$\times$ & 67.2\% & 90.6\% & 23.3\% & 36.6\% & \textcolor[HTML]{C00000}{43.9\%} & \textcolor[HTML]{C00000}{54\%}   \\
    15$\times$   & 56.8\% & 83\%   & 14.6\% & 19.5\% & \textcolor[HTML]{C00000}{42.2\%} & \textcolor[HTML]{C00000}{63.5\%} \\
    17.5$\times$ & 42.4\% & 81.3\% & 10.7\% & 13.3\% & \textcolor[HTML]{C00000}{31.7\%} & \textcolor[HTML]{C00000}{67\%}   \\
    \bottomrule
  \end{tabular}
  \vspace{-10pt}
\end{table}

\subsection{Quantifying the Semantic Priors ($F_{prior}$)}
\subsubsection{What is the quantitative impact of $F_{prior}$?}
To understand how the model compensates for degraded images, we quantified $F_{prior}$ by comparing the end-to-end generation probability against the zero-semantics dataset output. As compression increases and visual quality degrades, the model's reliance on $F_{prior}$ increases. As shown in Table \ref{tab:compression_results}, the semantic priors account for 23.8\% of the prediction at $7.5\times$ compression and 67\% at $17.5\times$ compression on the Fox dataset. The Omni dataset, which contains less predictable sequential context compared to Fox, shows a prior reliance fluctuating between 31.7\% and 45.3\%.

\subsubsection{Why is our framework the closest to the semantic vacuum?}
\label{Why is our framework the closest to the semantic vacuum}
To validate the semantic vacuum established by our Zero-Sense benchmark, we evaluate the posterior probability of the inserted text tokens conditioned on their surrounding context and demonstrate that the likelihood of a language model predicting these inserted words falls between $10^{-6}$ and $10^{-7}$. Across the dataset, no inserted token exhibits a prediction probability exceeding $10^{-5}$. 
In contrast, approaches for disrupting semantic coherence, such as token shuffling, permute the original text.
While such methods degrade local syntax, they fail to guarantee the eradication of incidental semantic correlations and leave residual predictive signals. By bounding the contextual predictability of our evaluation tokens to near-zero, our dataset construction framework provides a baseline, ensuring interpretability and validity when isolating visual retention.

\begin{table}[b]
  \caption{Decay of raw visual extraction capability ($OCR_{raw}$) across increasing compression ratios $\rho(\theta)$. The table reports the baseline performance isolated from the semantic priors, calibrated via the Zero-Sense benchmark. Results track the decline of $OCR_{raw}$ for the Fox and Omni datasets from $7.5\times$ to $17.5\times$ compression.}
  \label{tab:ocr_raw_results}
  \centering
  \renewcommand{\arraystretch}{1.2} 
  \setlength{\tabcolsep}{15pt} 
  \begin{tabular}{@{}ccc@{}} 
    \toprule
    \multirow{2}{*}{Compression} & \multicolumn{2}{c}{$OCR_{raw}$} \\
    \cmidrule(lr){2-3} 
          & Omni   & Fox    \\
    \midrule
    7.5$\times$  & 39.5\% & 76.1\% \\
    10$\times$   & 34\%   & 68.6\% \\
    12.5$\times$ & 28.5\% & 61\%   \\
    15$\times$   & 22.9\% & 53.5\% \\
    17.5$\times$ & 17.4\% & 46\%   \\
    \bottomrule
  \end{tabular}
  \vspace{-16pt}
\end{table}

\subsection{Quantitative Measurement of Raw Recognition ($OCR_{raw}$)}
\label{subsec:physical_bound}

To isolate $OCR_{raw}$ from the semantic priors, we evaluate performance following the calibration methodology outlined in \Cref{sec:Decoupling}. The decay of $OCR_{raw}$ across increasing compression ratios ($\rho(\theta)$) is recorded for the Omni and Fox datasets. As shown in Table \ref{tab:ocr_raw_results}, $OCR_{raw}$ declines from 39.5\% at 7.5$\times$ compression to 17.4\% at 17.5$\times$ compression for the Omni dataset. A degradation is observed on the Fox dataset, decreasing from 76.1\% to 46\% across the same interval.

\begin{table}[t]
  \centering
  \caption{Performance of DeepSeek-OCR under semantic interference with shuffled word orders. Herein, Original denotes natural document images, Shuffled denotes document images with shuffled word orders, and $\Delta$ (Drop) indicates the precision drop.} 
  \label{tab:ocr_compression_shuffled}
  \renewcommand{\arraystretch}{1.2}
  \setlength{\tabcolsep}{3pt}
  \begin{tabular}{@{}cccc|cccc@{}}
    \toprule
    \multirow{2}{*}{Compression} & \multicolumn{3}{c|}{Precision} & \multirow{2}{*}{Compression} & \multicolumn{3}{c}{Precision} \\
    \cmidrule(lr){2-4} \cmidrule(l){6-8}
    & Original & Shuffled & $\Delta$(Drop) & & Original & Shuffled & $\Delta$(Drop) \\
    \midrule
    2.5$\times$  & 97.4\% & 97.4\% & \textcolor[HTML]{C00000}{0.0\%}    & 12.5$\times$ & 88.0\% & 77.6\% & \textcolor[HTML]{C00000}{$-10.4\%$} \\
    5$\times$    & 97.0\% & 96.2\% & \textcolor[HTML]{C00000}{$-0.8\%$}  & 15$\times$   & 80.8\% & 66.1\% & \textcolor[HTML]{C00000}{$-14.7\%$} \\
    7.5$\times$  & 95.6\% & 92.5\% & \textcolor[HTML]{C00000}{$-3.1\%$}  & 17.5$\times$ & 70.5\% & 51.9\% & \textcolor[HTML]{C00000}{$-18.6\%$} \\
    10$\times$   & 92.8\% & 86.3\% & \textcolor[HTML]{C00000}{$-6.5\%$}  &              &        &        &                                     \\
    \bottomrule
  \end{tabular}
  \vspace{-6pt}
\end{table}

\subsection{Preliminary Perturbation Experiments}
\label{sec:simple_perturbation}

In addition to constructing semantically irrelevant data using a language model based text generator, we investigate alternative strategies that disrupt textual semantics. Prior studies systematically examine such approaches and categorize them into character level, token level, word level, and visual perturbations~\cite{chai2024tokenization,xing2025see,eger2019text}. These methods modify semantic content by manipulating symbolic units or visual glyphs, thereby producing texts with degraded semantics.

Guided by these perturbation strategies and the requirement to preserve the visual appearance of document images as much as possible, we conduct disordering experiments on the DDI-100 v1.3 dataset \cite{zharikov2020ddi}. This dataset provides document images together with two dimensional positional annotations for each word. Based on this information, we segment each word from the image and exchange word positions through size matching. The reconstructed images remain visually similar to the originals in terms of word size and layout, while the word order is altered.

We select 1478 images from DDI-100 v1.3 and perform five independent random permutations of word order for each image, resulting in five groups of shuffled images. Under the four resolution modes supported by DeepSeek-OCR, namely tiny, small, base, and large, we obtain OCR results for both the original and shuffled images. The average precision under different compression ratios is reported in Table \ref{tab:ocr_compression_shuffled}. Under the compression ratio of $17.5\times$ the precision gap is only 18.6\%, which indicates that exchanging word order introduces limited disruption to language level regularities and thus fail to satisfy the demands of our evaluation experiments.

\section{Conclusion}
In this paper, we investigated the existing evaluation protocols for visual-text compression methods and identified a critical oversight: the coupling of text preservation quality with the inherent linguistic priors of MLLMs. We argued that reliance on downstream task performance leads to biased assessments. To address this, we introduced a novel evaluation framework designed to decouple these confounding factors. The ZeroSense benchmark with low semantic correlation is further introduced to ensure that evaluation results purely reflect visual-text preservation. The extensive experiments reveal a significant divergence between VTC quality and downstream task accuracy. Our work not only offers a clearer understanding of current VTC methods but also establishes a more reliable foundation for the development of future long-context modeling architectures.

\clearpage

\bibliographystyle{unsrt}  
\bibliography{references}

\clearpage

\appendix

\section*{Supplementary Material}
\renewcommand\theHtable{Appendix.\thetable}
\renewcommand{\tablename}{Supplementary Table}
\setcounter{table}{0}

\appendix
\renewcommand{\theHfigure}{Appendix.\arabic{figure}} 
\setcounter{figure}{0}
\renewcommand{\figurename}{Supplementary Figure}

\section{Extended Evaluation Metrics: Normalized Edit Distance}
\label{sec:appendix_d}

To complement our measurement of $F_{prior}$, we additionally computed the edit distance metric to assist in assessing the impact of semantic priors on end-to-end outcomes. As shown in the following table, across compression ratios ranging from 7.5× to 17.5×, the model’s output performance is significantly compromised for both datasets after the elimination of semantic priors. This confirms that the model does not rely exclusively on text preservation, but rather leverages strong semantic priors to perform inference.

\begin{table}[htbp]
  \caption{Comparison of Edit Distance metrics with and without semantic relevance. Note: The negative sign denotes performance degradation under non-semantic conditions, not the absolute magnitude of the metric difference. This confirms that the model leverages strong semantic priors to perform inference.} 
  \label{tab:compression_results}
  \centering
  \renewcommand{\arraystretch}{1.2} 
  \setlength{\tabcolsep}{6pt} 
  \begin{tabular}{@{}ccccccc@{}}
    \toprule
    \multirow{2}{*}{Compression} & \multicolumn{2}{c}{$F_{prior} + OCR_{raw} \cdot K_{quality}$} & \multicolumn{2}{c}{$OCR_{raw} \cdot K_{quality}$} & \multicolumn{2}{c}{$F_{prior}$} \\
    \cmidrule(lr){2-3} \cmidrule(lr){4-5} \cmidrule(lr){6-7} 
          & Omni   & Fox    & Omni   & Fox    & Omni                   & Fox                   \\
    \midrule
    7.5$\times$  & 0.2 & 0.04 & 0.5 & 0.13 & \textcolor[HTML]{C00000}{-0.3}   & \textcolor[HTML]{C00000}{-0.09} \\
    10$\times$   & 0.21 & 0.07 & 0.58 & 0.22 & \textcolor[HTML]{C00000}{-0.37} & \textcolor[HTML]{C00000}{-0.15} \\
    12.5$\times$ & 0.34 & 0.06 & 0.63 & 0.32 & \textcolor[HTML]{C00000}{-0.29} & \textcolor[HTML]{C00000}{-0.26}   \\
    15$\times$   & 0.46 & 0.28   & 0.73 & 0.49 & \textcolor[HTML]{C00000}{-0.27} & \textcolor[HTML]{C00000}{-0.21} \\
    17.5$\times$ & 0.59 & 0.2 & 0.78 & 0.63 & \textcolor[HTML]{C00000}{-0.19} & \textcolor[HTML]{C00000}{-0.43}   \\
    \bottomrule
  \end{tabular}
  \vspace{-10pt}
\end{table}

\section{ZeroSense Details}

In this section, we provide supplementary visualizations and quantitative statistics to further validate the design and efficacy of the ZeroSense benchmark. First, we present qualitative comparisons between the original document images and their ZeroSense counterparts to illustrate our semantic decoupling process while preserving visual layout fidelity. Subsequently, we provide detailed quantitative statistics, including token posterior probability distributions and document token count distributions, to corroborate the empirical observations discussed in~\Cref{sec:Experiment}.

\begin{figure}[t]
    \centering
    \includegraphics[width=\linewidth]{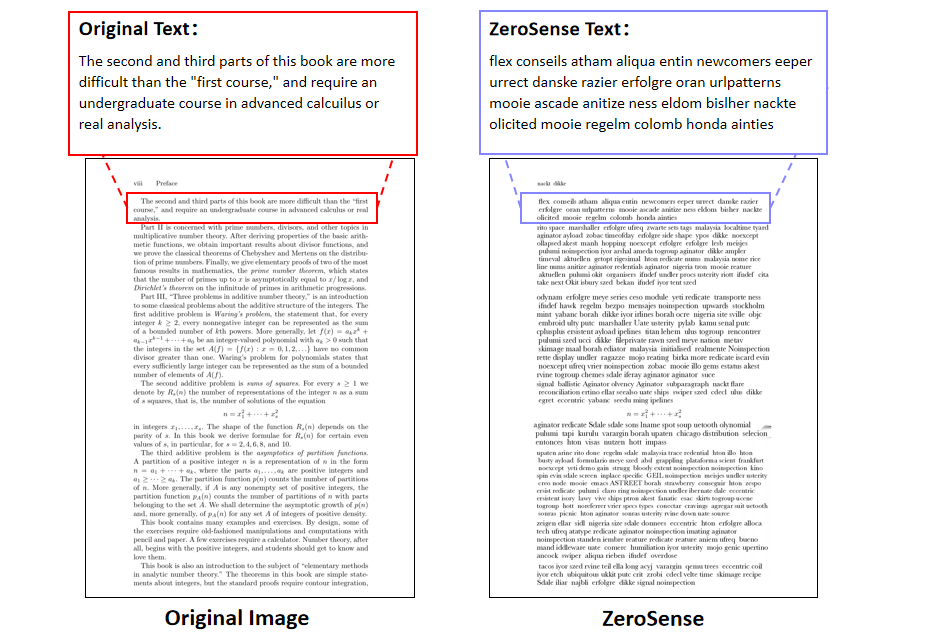}
    \caption{
   Comparison between original and ZeroSense document images. \textbf{Top:} Full page views demonstrate that our generation pipeline perfectly preserves the document's structural context. \textbf{Bottom:} Transcribed zoomed-in regions highlight the semantic decoupling; true semantic priors are systematically replaced with tokens sampled from a low-probability vocabulary subset, isolating the visual layout characteristics.
    }
    \label{fig:comparison2}
\end{figure}


\subsection{Visualizing the ZeroSense}
To demonstrate the semantic decoupling achieved by the ZeroSense benchmark,we provide a comparison of original document images and their rendered counterparts (see Supplementary~\Cref{fig:comparison2}). While original datasets contain linguistic priors , the ZeroSense generation pipeline systematically replaces these with tokens sampled from a low-probability vocabulary subset. Detailed quantitative results will be presented in the next section.

\subsection{Data Statistics of ZeroSense}
\begin{figure}[htbp]  
    \centering  
    \begin{subfigure}[b]{0.45\linewidth}
        \centering
        \includegraphics[width=\linewidth]{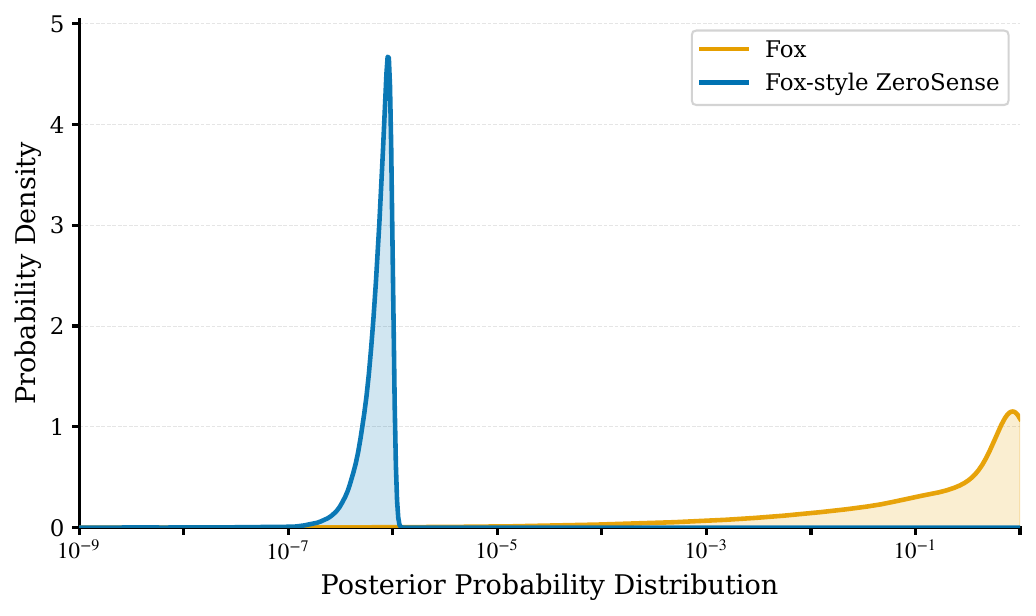}
        \caption{Fox and Foxstyle-ZeroSense}  
        \label{fig:fox_prob}     
    \end{subfigure}
    \hfill  
    \begin{subfigure}[b]{0.45\linewidth}
        \centering
        \includegraphics[width=\linewidth]{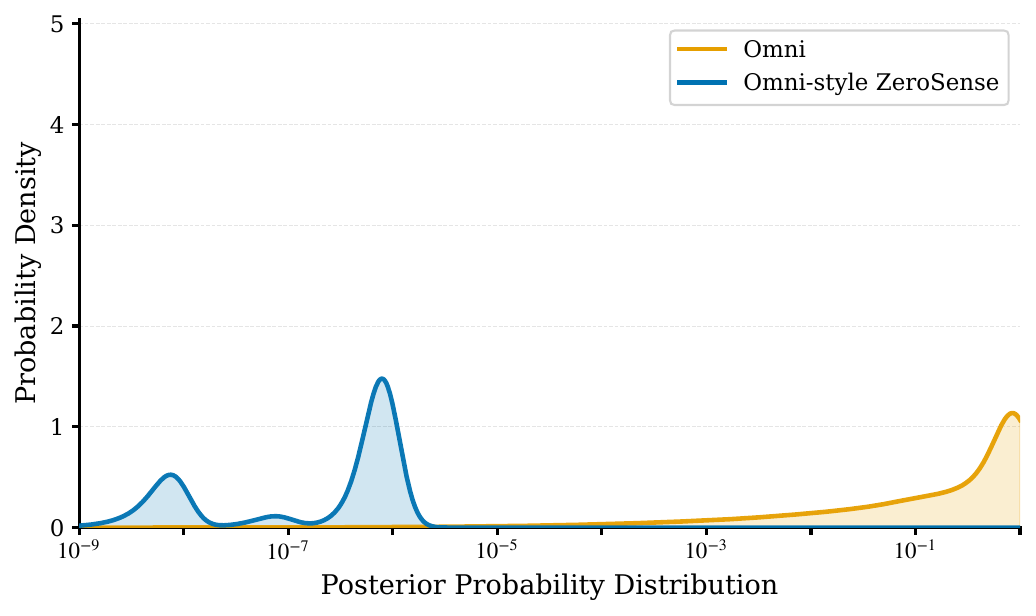}
        \caption{Omni and Omnistyle-ZeroSense}  
        \label{fig:omni_prob}     
    \end{subfigure}
    
    \caption{We compared the posterior probability distribution plots of textual content between Fox, Omni, and their corresponding rendered ZeroSense data.
Yellow denotes the original data, and blue represents the rendered data.
It can be clearly observed that the text in the original dataset exhibits strong semantic priors, whereas the textual content in ZeroSense is highly semantically irrelevant.}
    \label{fig:probabilities_all}
\end{figure}

To verify the semantic priors in~\Cref{sec:Semantically irrelevant sample generation} and~\Cref{Why is our framework the closest to the semantic vacuum}, we visualize the distribution of token posterior probabilities. Supplementary Figure \ref{fig:probabilities_all} shows the detailed posterior probability distributions, demonstrating the reliability and interpretability of ZeroSense’s contextual semantic independence.

To clarify the characteristic differences between Omni and Fox in~\Cref{subsec:main_results}, we provide statistical results on their token count distributions.
Omni has a considerable fraction of extreme cases, while Fox shows a more reasonable token distribution. This aligns with the reasoning presented in~\Cref{subsec:main_results} for why, under the distinct rendering schemes of the two datasets and at a compression ratio of 17.5×, the precision difference between Fox and Foxstyle‑ZeroSense is larger than that relative to Omni.
The rendering scheme of Omni involves a large number of extremely distributed cases that nearly exceed the OCR recognition capability of the model, which cannot be corrected even on the basis of semantic priors.
Consequently, the discrepancy between the extracted real text preservation rate and the end-to-end value is relatively small. 
\begin{figure}
    \centering
    \includegraphics[width=1\linewidth]{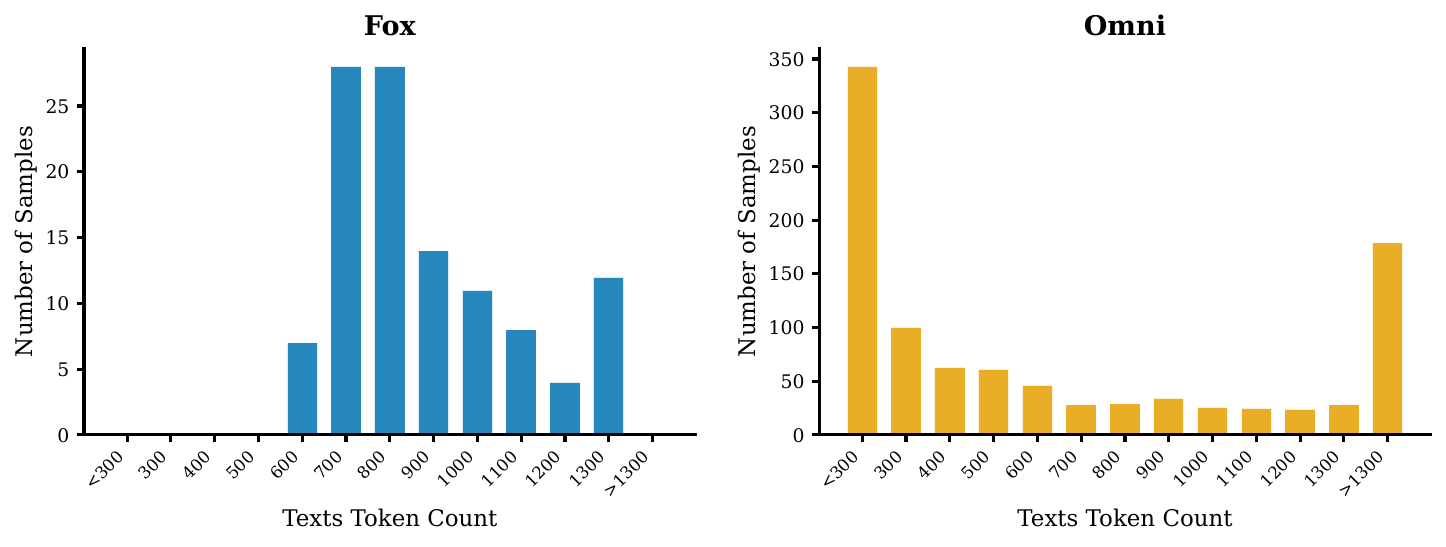}
    \caption{The figures respectively illustrate the distribution of the number of text tokens for Fox and Omni. Fox is distributed almost entirely within a reasonable range, whereas Omni lies mostly in extreme regions.
This aligns with the results presented in~\Cref{subsec:main_results}. Owing to the extreme nature of Omni’s rendering scheme, after disentangling semantic priors and OCR recognition accuracy, the difference in text preservation between OmniStyle-ZeroSense and Omni is smaller than that between FoxStyle-ZeroSense and Fox.}
    \label{fig:placeholder}
\end{figure}

\section{Algorithm Details and Pseudocode of ZeroSense Construction}
\label{sec:appendix_c}

To ensure the reproducibility of the ZeroSense benchmark generation, this section details the layout feature extraction and semantic generation pipeline discussed in~\Cref{sec:benchmark}. Depending on the annotation quality and background complexity of the source dataset, our framework applies adaptive engineering strategies. For documents with clean backgrounds (e.g., Fox), we render the extracted attributes onto a normalized blank canvas to eliminate scaling artifacts. For datasets with complex layouts and backgrounds (e.g., Omni), we apply an auxiliary inpainting algorithm to erase the original text prior to rendering.

\subsection{Bottom-up Layout Reconstruction}

For datasets lacking render-friendly paragraph-level positional information, we employ a bottom-up layout reconstruction algorithm (Algorithm \ref{alg:layout_reconstruction}). This extraction algorithm aligns with the process described in~\Cref{sec:Feature_Extraction}. It takes a set of raw word-level OCR bounding boxes $\mathcal{B}_{raw}$ as input. Based on the given coordinates $B_{pos} = (x, y, w, h)$, each box $b \in \mathcal{B}_{raw}$ is parameterized mathematically by its left $x$-coordinate ($x_b$), width ($w_b$), height ($h_b$), and center $y$-coordinate ($y^c_b$). The algorithm relies on vertical projection profiles to identify column gaps. To mitigate noise, we introduce a noise tolerance threshold $T_{noise} = 3.0 \times H_{unified}$ (where $H_{unified}$ is the median line height). Scattered boxes are heuristically merged into paragraph-level blocks using a spatial distance function $\mathrm{HorizontalGap}(\cdot, \cdot)$ and a bounding box union operator $\mathrm{Merge}(\cdot, \cdot)$.

\begin{algorithm}
\caption{Bottom-up Bounding Box Reconstruction}
\label{alg:layout_reconstruction}
\begin{algorithmic}[1]
\REQUIRE Raw OCR bounding box set $\mathcal{B}_{raw}$, Image Width $W$. 
\REQUIRE Helper functions: $\mathrm{HorizontalGap}(u, v)$ yields horizontal distance, $\mathrm{Merge}(u, v)$ yields the geometric union box, $\mathrm{Pop}(\mathbb{S})$ extracts and removes an element from set $\mathbb{S}$.
\ENSURE Reconstructed paragraph-level bounding boxes $\mathcal{B}_{merged}$

\STATE \textbf{Initialize:} $H_{unified} \leftarrow \mathrm{median}(\{h_b \mid b \in \mathcal{B}_{raw} \textbf{ and } h_b > 8\})$
\STATE \textbf{Initialize:} Projection array $P \leftarrow \text{Zeros}(W)$

\STATE \textit{// Step 1: Vertical Projection with Noise Filtering}
\FOR{\textbf{each} $b \in \mathcal{B}_{raw}$}
    \FOR{$x \leftarrow x_b \textbf{ to } x_b + w_b$}
        \STATE $P[x] \leftarrow P[x] + h_b$ \hfill // \textit{Accumulate text height density}
    \ENDFOR
\ENDFOR

\STATE \textit{// Step 2: Column Splitting}
\STATE $T_{noise} \leftarrow H_{unified} \times 3.0$
\STATE $G_{min} \leftarrow H_{unified} \times 0.8$ \hfill // \textit{Minimum horizontal gap to split columns}
\STATE Partition $\mathcal{B}_{raw}$ into independent column sets $\mathcal{C} = \{C_{1}, \dots, C_{n}\}$ by finding horizontal gaps where $P[x] < T_{noise}$ for a continuous width $\ge G_{min}$.

\STATE \textit{// Step 3: Heuristic Neighborhood Aggregation}
\STATE $\mathcal{B}_{merged} \leftarrow \emptyset$
\FOR{\textbf{each} column $C_{i} \in \mathcal{C}$}
    \WHILE{$C_{i} \neq \emptyset$}
        \STATE $u \leftarrow \mathrm{Pop}(C_{i})$ \hfill // \textit{Extract current reference box}
        \FOR{\textbf{each} remaining box $v \in C_{i}$}
            \STATE $\Delta y \leftarrow |y^c_u - y^c_v|$
            \STATE $\Delta x \leftarrow \mathrm{HorizontalGap}(u, v)$
            \IF{$\Delta y < H_{unified} \times 0.5 \textbf{ and } \Delta x < H_{unified} \times 1.5$}
                \STATE $u \leftarrow \mathrm{Merge}(u, v)$
                \STATE \text{Remove} $v$ \text{from} $C_{i}$
            \ENDIF
        \ENDFOR
        \STATE \text{Add} $u$ \text{to} $\mathcal{B}_{merged}$
    \ENDWHILE
\ENDFOR
\RETURN $\mathcal{B}_{merged}$
\end{algorithmic}
\end{algorithm}

\subsection{Geometry-Constrained Font Size Solver}

Once the bounding box dimensions $(w, h)$ and textual content $T_i$ for a text block $B_i$ are established, we determine the precise layout parameters. Algorithm \ref{alg:font_solver} outlines the implementation of the geometry-constrained inverse solver defined in~\Cref{eq:max}, alongside the language-aware capacity calculation $C_i$ defined in~\Cref{eq:count}. We utilize the language detection function $L(\cdot)$ and the rendering simulation function $f(T_i, s)$ to calculate the physical dimensions $w(f(T_i,s))$ and $h(f(T_i,s))$ of the text at a given font size $s$. A language-specific buffer coefficient $\beta$ is applied to approximate $w(f(T_i,s)) \approx w$.

\begin{algorithm}[htbp]
\caption{Geometry-Constrained Inverse Solver and Capacity Calculation}
\label{alg:font_solver}
\begin{algorithmic}[1]
\REQUIRE Bounding box width $w$ and height $h$, Text block $T_i$
\REQUIRE Helper functions: $L(\cdot)$ detects language, $\mathrm{count_{word}}(\cdot)$, $\mathrm{count_{char}}(\cdot)$, $f(T_i, s)$ simulates rendering
\ENSURE Optimal font size $S_{opt}$, Text Capacity $C_i$

\STATE \textit{// Step 1: Language-aware Capacity Calculation (Enforce~\Cref{eq:count})}
\IF{$L(T_i) = \text{Latin}$}
    \STATE $C_i \leftarrow \mathrm{count_{word}}(T_i)$
    \STATE $\beta \leftarrow 1.05$ \COMMENT{Buffer coefficient for Latin scripts}
\ELSE
    \STATE $C_i \leftarrow \mathrm{count_{char}}(T_i)$
    \STATE $\beta \leftarrow 1.01$ \COMMENT{Buffer coefficient for Logographic systems}
\ENDIF

\STATE \textit{// Step 2: Iterative Font Size Search (Enforce~\Cref{eq:max})}
\STATE $S_{start} \leftarrow \min(h, 100)$
\STATE $S_{opt} \leftarrow 8$ \COMMENT{Minimum fallback font size}

\FOR{$s \leftarrow S_{start}$ \textbf{down to} 8}
    \STATE $w_{sim} \leftarrow w(f(T_i, s))$ \COMMENT{Simulated physical width}
    \STATE $h_{sim} \leftarrow h(f(T_i, s))$ \COMMENT{Simulated physical height}
    
    \STATE \textit{// Calculate required vertical space constrained by bounding box width $w$}
    \STATE $N_{req} \leftarrow \max(1, \lfloor (w_{sim} \times \beta) / w \rfloor + 1)$
    \STATE $H_{req} \leftarrow N_{req} \times h_{sim}$
    
    \IF{$H_{req} \le h$}
        \STATE $S_{opt} \leftarrow s$
        \STATE \textbf{break}
    \ENDIF
\ENDFOR

\RETURN $S_{opt}, C_i$
\end{algorithmic}
\end{algorithm}

\subsection{Semantically Irrelevant Sample Generation}

To construct the ZeroSense benchmark, we generate semantically irrelevant textual content to replace the original document text $T$. Algorithm \ref{alg:text_generation} details this replacement algorithm, which operationalizes the formulation in ~\Cref{eq:valid} to disrupt contextual predictability. 

We utilize Qwen2.5-7B as our base causal language model. During the generation loop, the model acts purely as a probability evaluator. For each step $t$, we force the model to sample the next token $w$ from a restricted vocabulary subset $Q_{t}$ defined by a low-probability threshold $\tau$ (initialized to $\tau_{init}$), given the context $x_{<t}$. If the restricted vocabulary subset $Q_t$ becomes empty, the algorithm applies a dynamic threshold relaxation strategy to ensure generation robustness while maintaining the semantic vacuum.

\begin{algorithm}[htbp]
\caption{Causal LM-based Semantically Irrelevant Text Generation}
\label{alg:text_generation}
\begin{algorithmic}[1]
\REQUIRE Target capacity $C_{i}$, Initial low-probability threshold $\tau_{init}$, Valid vocabulary space $V_{valid}$
\REQUIRE Helper function: $\mathrm{UniformSample}(\mathbb{S})$ samples one element uniformly from set $\mathbb{S}$
\ENSURE Semantically irrelevant token sequence $T_{zero}$

\STATE \textbf{Initialize:} Context sequence $x_{<1} \leftarrow [\mathrm{BOS}]$
\STATE \textbf{Initialize:} Generated sequence $T_{zero} \leftarrow \emptyset$

\FOR{$t \leftarrow 1$ \textbf{to} $C_{i}$}
    \STATE \textit{// 1. Obtain posterior probability distribution}
    \STATE Compute next-token probabilities $P(\cdot \mid x_{<t})$ using the causal LM
    
    \STATE \textit{// 2. Dynamic threshold filtering (Enforce~\Cref{eq:valid})}
    \STATE $\tau \leftarrow \tau_{init}$
    \STATE $Q_{t} \leftarrow \emptyset$
    \STATE $attempts \leftarrow 0$
    
    \WHILE{$Q_{t} = \emptyset \wedge attempts < 10$}
        \STATE $Q_{t} \leftarrow \{ w \in V_{valid} \mid P(w \mid x_{<t}) < \tau \}$ \COMMENT{Restricted vocabulary subset}
        
        \IF{$Q_{t} = \emptyset$}
            \STATE $\tau \leftarrow \tau \times 10$ \COMMENT{Dynamic threshold relaxation}
            \STATE $attempts \leftarrow attempts + 1$
        \ENDIF
    \ENDWHILE
    
    \STATE \textit{// 3. Token Selection and Context Update}
    \IF{$Q_{t} = \emptyset$}
        \STATE $w_{t} \leftarrow \mathrm{UniformSample}(V_{valid})$ \COMMENT{Fallback to prior space}
    \ELSE
        \STATE $w_{t} \leftarrow \mathrm{UniformSample}(Q_{t})$
    \ENDIF
    
    \STATE $x_{<t+1} \leftarrow x_{<t} \oplus [w_{t}]$ \COMMENT{Append $w_t$ to context sequence}
    \STATE $T_{zero} \leftarrow T_{zero} \cup \{w_{t}\}$
\ENDFOR

\RETURN $T_{zero}$
\end{algorithmic}
\end{algorithm}

\clearpage

\section{Preliminary Perturbation Experiments Details}
 This section details the  construction pipeline and the case of preliminary perturbation dataset discussed in~\Cref{sec:simple_perturbation}. 
\subsection{The Pipeline of the Perturbed Dataset Construction}

Our perturbation strategy requires a document image dataset with bounding box annotations, and the data source we finally adopt is DDI-100 v1.3, which provides document images along with the 2D coordinates of each word's position to enable effective word cropping and rearrangement. Through this construction process, the semantic content of the document is altered, while the original visual layout and pixel-level statistical properties are largely preserved. As illustrated in Supplementary~\Cref{fig:pipeline}, the perturbation procedure consists of two stages:

\textbf{(1) Geometry-constrained inter-line rearrangement}, where all text lines are initially grouped based on geometric constraints, with the size tolerance of lines in the same group constrained within 5\%, and only lines within the same group are permitted to exchange positions randomly;

\textbf{(2) Intra-line word shuffling with preserved visual density}, which further disrupts semantic coherence by randomly reordering the words within each line. The original visual density is maintained by constraining the spacing between words after reordering to be consistent with that before reordering. Ultimately, the constructed shuffled images closely replicate the visual appearance of the original images.

\begin{figure}[htbp]
    \centering
    \includegraphics[width=0.9\linewidth]{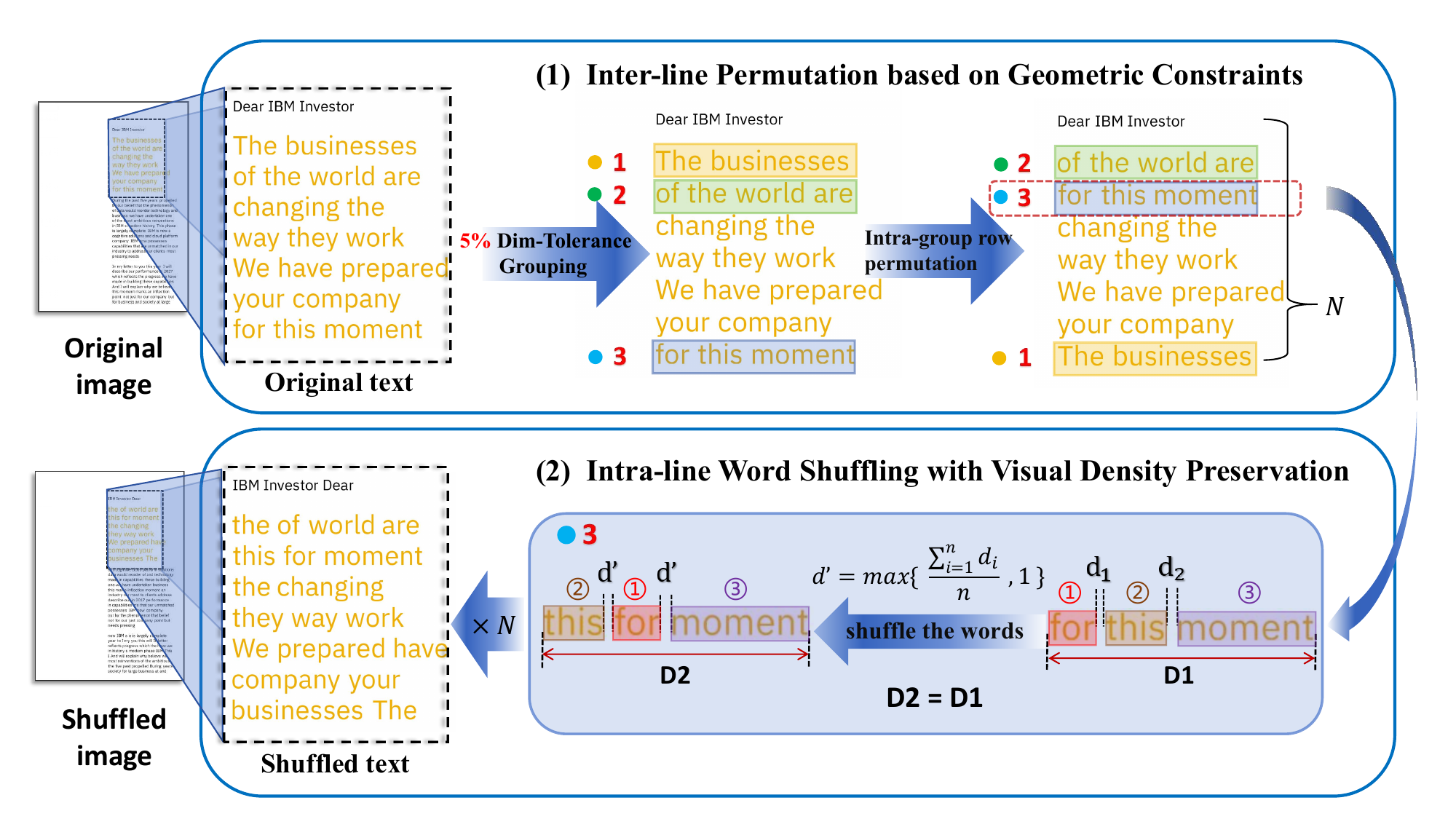}
    \caption{
    Illustration of the two-stage shuffling process applied to an original document image: (1) Inter-line Permutation based on Geometric Constraints and (2) Intra-line Word Shuffling with Visual Density Preservation.
    }
    \label{fig:pipeline}
\end{figure}

\subsection{Case of the Perturbed Dataset}
\begin{figure}[htbp]
    \centering
    \includegraphics[width=\linewidth]{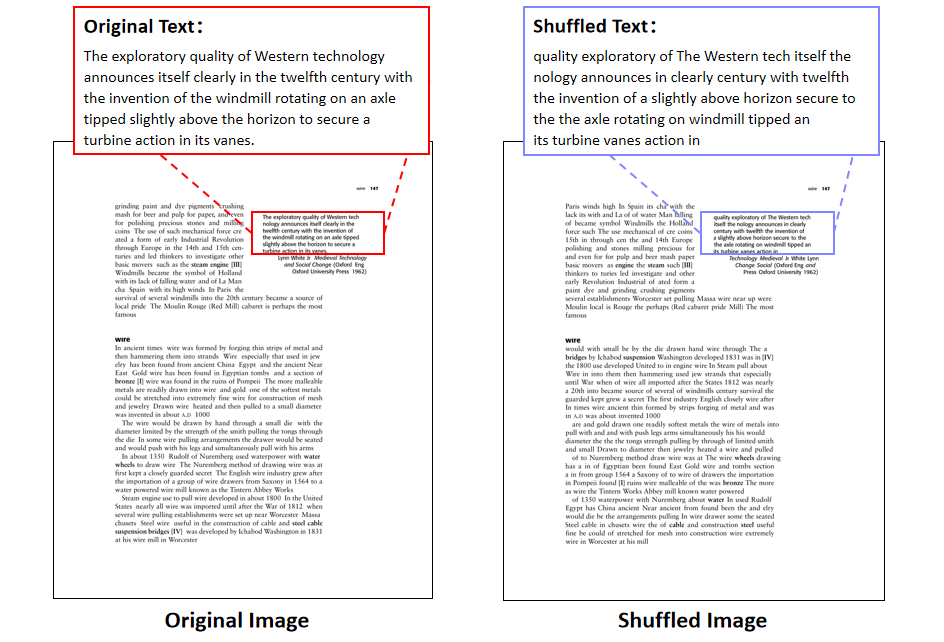}
    \caption{
    Comparison between original and shuffled document images. Dislike the original image, the shuffled image exhibits a scrambled word sequence.
    }
    \label{fig:comparison}
\end{figure}

Supplementary~\Cref{fig:comparison} presents a representative case from the perturbed dataset, displaying both the original image with intact word order and the resulting shuffled image. Although the shuffled image exhibits a scrambled word sequence, visual characteristics such as layout and font size remain nearly identical to the original image.

\end{document}